\pdfoutput=1 
\documentclass[letterpaper]{article} 
\usepackage[preprint]{aaai2027} 
\usepackage[hyphens]{url}
\usepackage{graphicx}
\usepackage{natbib}
\usepackage{caption}
\usepackage{amsmath}
\usepackage{amssymb}
\usepackage{mathtools}
\usepackage{booktabs}
\usepackage{algorithm}
\usepackage{algorithmic}

\newcommand{\tabfont}{\fontsize{9}{10.5}\selectfont}

\title{EPI-SST: Renewal-Structured State-Space Transformers for Epidemic Forecasting}
\author{
Yiyun Chen\textsuperscript{\rm 1},
Zewen Liu\textsuperscript{\rm 1},
Shengpu Tang\textsuperscript{\rm 1},
Max S. Y. Lau\textsuperscript{\rm 2},
Wei Jin\textsuperscript{\rm 1}\corresponding
}
\affiliations{
\textsuperscript{\rm 1}Department of Computer Science, Emory University, Atlanta, GA, USA\\
\textsuperscript{\rm 2}Department of Biostatistics and Bioinformatics, Emory University, Atlanta, GA, USA\\
yiyun.chen@emory.edu, zewen.liu@emory.edu, shengpu.tang@emory.edu,\\
msy.lau@emory.edu, wei.jin@emory.edu
}
\begin{document}
\maketitle
\begin{abstract}
Epidemic forecasting supports public health decisions ranging from capacity planning to intervention timing, for which both point accuracy and calibrated uncertainty are essential. Although neural models have advanced point forecasting, producing reliable probabilistic predictions over multiple future horizons remains challenging. Renewal models encode epidemic propagation, while neural forecasters learn flexible multi-horizon representations; integrating these strengths within a probabilistic future rollout remains underexplored. We propose EPI-SST (Epidemic State Space Transformer), a probabilistic forecaster centered on an evidence-driven renewal transition. At each horizon, neural evidence produces a bounded update to an internal multiplier and advances a finite-memory renewal state that generates the future mean trajectory. A separate heteroscedastic Gaussian head controls predictive scale using a validation-selected log-variance penalty. Across four epidemic surveillance benchmarks, EPI-SST achieves the lowest pooled RMSE on every dataset, reducing RMSE by up to 14.2\% relative to the strongest baseline. It also achieves the lowest continuous ranked probability score (CRPS) at horizons \(h=4,5,6\) on every dataset, with reductions of up to 43\% relative to the strongest probabilistic baseline. During the abnormal 2019--2020 influenza season, EPI-SST reduces CRPS by 16.9\% and calibration error by 78.9\%; a three-seed replication preserves its distributional lead. Ablation studies and frozen-path interventions show that the renewal transition, attention conditioning, horizon coupling, and uncertainty-scale control make complementary contributions to predictive performance.
\end{abstract}

\section{Introduction}

Epidemic forecasting informs concrete operational decisions, including how
much hospital capacity to reserve, when to activate overflow wards, and when
to tighten or relax non-pharmaceutical interventions~\citep{liu2024review}.
During the 2020--21 COVID winter surge, hospital administrators had to decide
weeks in advance whether to cancel elective procedures. Rising point
forecasts could indicate an impending surge, but without calibrated
uncertainty bands, they provided no principled threshold for initiating
contingency plans~\citep{cramer2022covid}. Multiyear evaluations of influenza
and COVID-19 forecasts~\citep{reich2019flu,cramer2022covid}, together with
proper scoring frameworks~\citep{gneiting2007proper}, have established that
operational value depends on the quality of the full predictive distribution.
Accordingly, epidemic forecasting benchmarks now report CRPS, calibration
diagnostics, and interval coverage alongside RMSE~\citep{bracher2021wis}.

Existing approaches address complementary aspects of this problem but do not
jointly provide epidemiologically structured multi-horizon forecasts and
explicit uncertainty control. Mechanistic and semi-mechanistic renewal models
offer explicit transmission structure and principled uncertainty
quantification~\citep{hethcote2000mathematics,cori2013new,abbott2020estimating}.
However, multi-step forecasting requires future transmission dynamics to be
specified or extrapolated, and inference is commonly performed separately for
individual datasets or series rather than amortized across regions. Neural
forecasters~\citep{adhikari2019epideep,zhu2019attention,shahid2020predictions,nie2022patchtst}
support amortized forecasting, while hybrid approaches incorporate
epidemiological information through training objectives, latent
representations, or architectural
constraints~\citep{rodriguez2022einn,wang2020tdefsi,sun2025epi}. These
approaches, however, do not use a renewal recursion as the explicit transition
governing the evolution of the multi-horizon predictive mean, and most do not
produce a predictive distribution. Probabilistic neural forecasters provide
expressive predictive distributions but differ in the extent to which they
couple forecast horizons. EpiFNP~\citep{kamarthi2021epifnp} fits a separate
model for each horizon, leaving its predicted distributions uncoupled across
leads. RNP~\citep{qin2019rnp} produces all horizons using a single model, but
does not constrain how the predictive mean evolves from one horizon to the
next.

Latent state-space models
(SSMs)~\citep{rangapuram2018deepssm,tang2021protran,bao2023usst} provide a
natural framework for addressing horizon coupling. By propagating one latent
state at each forecast step, they establish an explicit dependency across
successive leads. Nevertheless, two limitations remain in the epidemic
forecasting setting. First, their transition dynamics are designed for
generic sequence modeling rather than constrained by an epidemic renewal
operator. Consequently, the predictive mean is not required to evolve
multiplicatively through a lag-weighted incidence history, despite this being
the mechanism through which incidence propagates. The resulting rollout
therefore lacks explicit epidemic growth structure and remains free to drift
at longer horizons or under regime shifts. Second, their training objectives
place no explicit constraint on predictive scale. Under a Gaussian
likelihood, the predicted variance can absorb errors in the mean by expanding
around isolated large residuals rather than representing systematic forecast
uncertainty. We therefore identify two central challenges:
\emph{structured horizon coupling} and \emph{uncertainty-scale control}.

\begin{figure*}[t]
\centering
\includegraphics[width=0.98\textwidth]{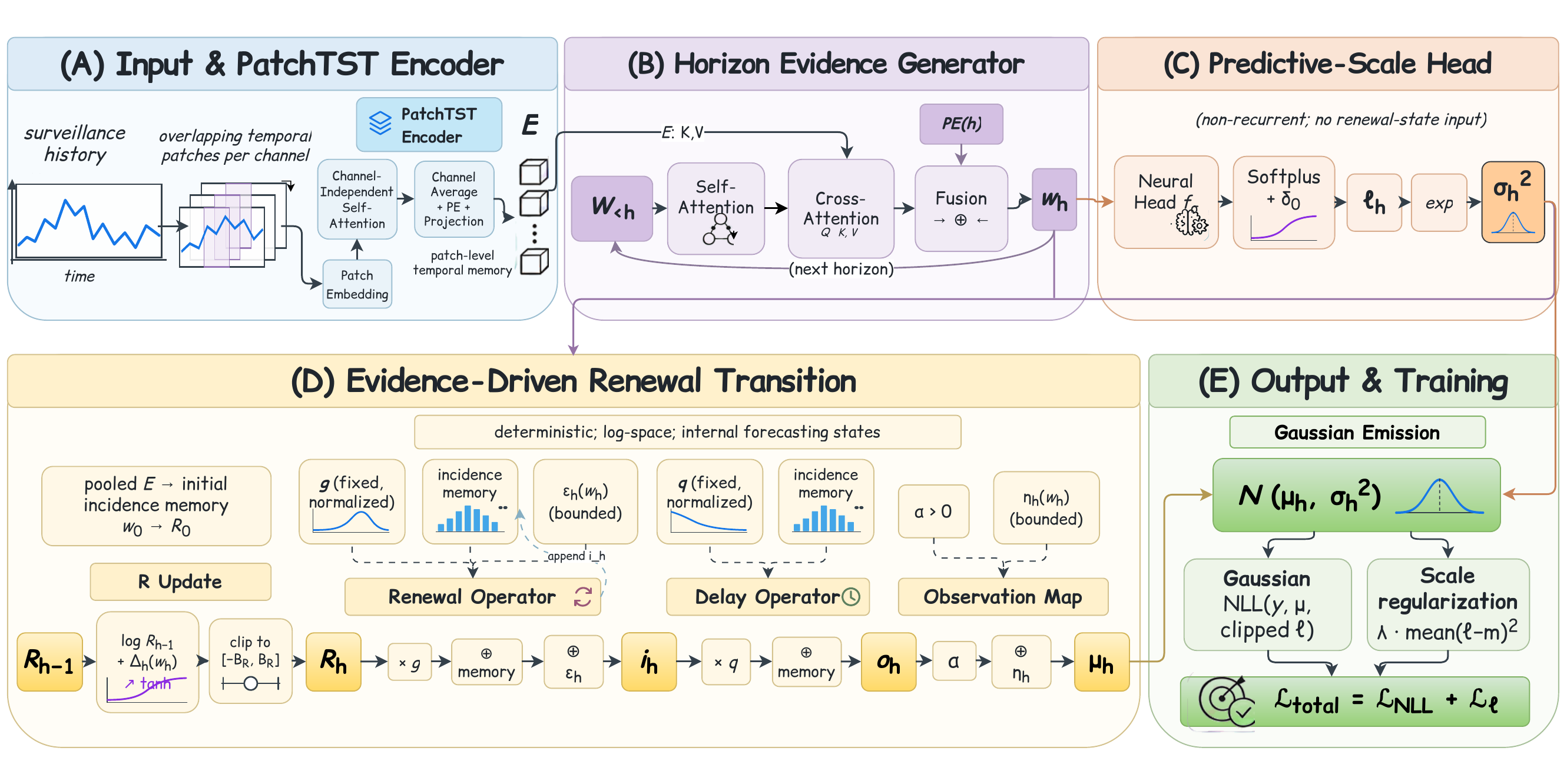}
\caption{EPI-SST. \textbf{(A)} PatchTST encodes surveillance history into
memory $\mathbf{E}$. \textbf{(B)} Self-attention over earlier evidence,
cross-attention to $\mathbf{E}$, and a horizon embedding produce
$\mathbf{w}_h$. The shared evidence then branches into two paths:
\textbf{(C)} a direct non-recurrent head maps $\mathbf{w}_h$ to predictive
variance $\sigma_h^2$, while \textbf{(D)} bounded controls update $R_h$ and
fixed normalized lag kernels propagate incidence $i_h$ to delayed signal
$o_h$ and predictive mean $\mu_h$. \textbf{(E)} Mean and variance meet only
at the Gaussian emission; training combines Gaussian NLL with the
validation-selected log-variance penalty toward reference $m$.}
\label{fig:model_overview}
\end{figure*}

Inspired by latent state-space models, we propose EPI-SST (Epidemic State
Space Transformer), a novel framework that replaces the generic latent
transition with a finite-memory renewal transition and estimates predictive
scale along a separate path (Figure~\ref{fig:model_overview}). To address
structured horizon coupling, a PatchTST encoder~\citep{nie2022patchtst}
produces horizon-conditioned evidence that drives bounded, clipped updates to
a model-internal multiplier $R_h$. Fixed normalized lag kernels then propagate
latent incidence in log space to a delayed signal and subsequently to the
predictive mean. In this way, EPI-SST incorporates the renewal
equation~\citep{fraser2007estimating,wallinga2006generation} as a forward
generation rule rather than a retrospective estimator. The imposed structure
is intentionally minimal, consisting of one multiplicative recursion and two
lag kernels that are shared unchanged across all four diseases. It therefore
serves as an inductive bias rather than a complete generative specification.
To address uncertainty-scale control, a separate heteroscedastic head maps the
same horizon-conditioned evidence to predictive variance under a
validation-selected log-variance penalty. This design controls predictive
scale without allowing it to enter the renewal state. We summarize our
contributions as follows.

\begin{itemize}
\item We develop EPI-SST, a novel probabilistic forecasting framework that
uses the renewal equation as a forward transition within a learned
multi-horizon rollout and estimates uncertainty along a separate path outside
the recursion. This design distinguishes EPI-SST from renewal-based
reconstruction and latent state-space forecasters with unconstrained
transitions.

\item We evaluate EPI-SST on four surveillance benchmarks spanning continuous
and sparse count targets. It achieves the lowest pooled RMSE on every dataset,
reducing RMSE by up to $14.2\%$ relative to the strongest baseline. At
$h{=}4,5,6$, it achieves the lowest CRPS on every dataset, improving over the
strongest probabilistic baseline by $8$--$43\%$. The distributional gains span
continuous and sparse targets and persist during the abnormal 2019--2020
influenza season.

\item Frozen-checkpoint interventions show that the learned renewal path
contributes functionally to prediction. Reversing the recovered $R_{1:H}$
trajectory while preserving its level and values increases CRPS by $14.3\%$.
Starting from paths exchanged across forecast windows, progressively restoring
each sample's original shape recovers pooled RMSE and CRPS. Perturbing $R_1$
also shifts forecasts in the expected direction with no sign violations.
\end{itemize}

\section{Related Work}

EPI-SST draws on latent state-space models for horizon coupling, epidemic
forecasters for the target task, and renewal-process models for its structured
transition.

\noindent\textbf{Latent State-Space Models.}
Latent state-space models combine neural representations with parameterized
state transitions. DSSM predicts the parameters of a linear Gaussian
state-space model~\citep{rangapuram2018deepssm}; ProTran introduces
step-specific latent variables into a probabilistic
Transformer~\citep{tang2021protran}; USST extends the state-space Transformer
to uncertainty-aware forecasting~\citep{bao2023usst}; and S4 structures the
recurrence for long sequences~\citep{gu2022s4}. All couple forecast steps
through a propagated latent state, a property retained by EPI-SST. The
principal distinction lies in the transition and predictive scale: these
models use generic sequence transitions and learn uncertainty through their
native probabilistic objectives, whereas EPI-SST imposes a renewal operator
and adds an explicit validation-selected reference-scale constraint.
Conformal and Gaussian-process approaches pursue calibrated
intervals through coverage-oriented procedures rather than transition
constraints~\citep{stankeviciute2021conformalts,papadopoulos2024guaranteed}.

\noindent\textbf{Epidemic Forecasting Models.}
RNP~\citep{qin2019rnp} and EpiFNP~\citep{kamarthi2021epifnp} are the closest
learned probabilistic baselines, while regional operational systems emphasize
calibrated decision support~\citep{ravenda2025probabilistic,engebretsen2023realtime}.
A complementary line incorporates epidemiological structure into
architectures, representations, or objectives, including
EpiDeep~\citep{adhikari2019epideep}, EINN~\citep{rodriguez2022einn},
TDEFSI~\citep{wang2020tdefsi}, and Epi$^2$-Net~\citep{sun2025epi}. EPI-SST
differs in where this structure acts: the renewal equation defines the
transition generating the multi-horizon mean rather than acting only through
the encoder or training objective.

\noindent\textbf{Renewal-Process Models of Transmission.}
The renewal equation relates incident infections to a time-varying reproduction number and a generation-interval-weighted history~\citep{fraser2007estimating,wallinga2006generation}. It underlies branching-process estimators~\citep{wallinga2004different,cori2013new}, semi-mechanistic Bayesian models of delayed outcomes~\citep{flaxman2020estimating,bhatt2023semi}, and tools such as EpiNow2~\citep{abbott2020estimating}. EPI-SST runs the renewal recursion prospectively, using encoded history to generate future multipliers within a transition shared across regions.

\section{Methods}
\label{sec:methods}

Given the normalized surveillance history
$\mathbf{X}_{t_0}=\mathbf{x}_{t_0-L+1:t_0}\in\mathbb{R}^{L}$ ending at
forecast origin $t_0$, EPI-SST produces $H$ horizon-specific Gaussian
forecasts by rolling a finite-memory renewal state forward one step at a
time. A PatchTST encoder supplies horizon-conditioned evidence for each
update: the renewal branch generates the predictive mean $\mu_h$, while a
separate emission head predicts the variance $\sigma_h^2$. The conditional
forecast factorizes as
$p_{\theta}(\mathbf{y}_{t_0+1:t_0+H}\mid\mathbf{X}_{t_0})
=\prod_{h=1}^{H}\mathcal{N}(y_{t_0+h};\mu_h,\sigma_h^{2})$.
The two branches share the same evidence sequence, but only the renewal
branch advances the state that couples successive forecast horizons. Each
horizon can therefore retrieve information from the observed history, while
dependence among future means is carried by repeated application of the same
state transition rather than by independently decoded outputs.
Figure~\ref{fig:model_overview} summarizes the complete framework.

\begin{table*}[!t]
\centering
\begingroup
\tabfont
\setlength{\tabcolsep}{4.2pt}
\begin{tabular}{@{}lcccccccc@{}}
\toprule
& \multicolumn{2}{c}{COVID Hospital Adm.} & \multicolumn{2}{c}{COVID Deaths} & \multicolumn{2}{c}{Measles} & \multicolumn{2}{c}{Mumps} \\
\cmidrule(lr){2-3} \cmidrule(lr){4-5} \cmidrule(lr){6-7} \cmidrule(lr){8-9}
Model & RMSE & MAE & RMSE & MAE & RMSE & MAE & RMSE & MAE \\
\midrule
CNN      & $0.402{\pm}.023$ & $0.271{\pm}.026$ & $0.376{\pm}.010$ & $0.242{\pm}.021$ & $0.458{\pm}.082$ & $0.450{\pm}.084$ & $0.354{\pm}.032$ & $0.344{\pm}.032$ \\
DLinear  & $0.473{\pm}.010$ & $0.385{\pm}.014$ & $0.463{\pm}.011$ & $0.367{\pm}.012$ & $0.229{\pm}.014$ & $0.210{\pm}.013$ & $0.135{\pm}.009$ & $0.112{\pm}.010$ \\
GRU      & $0.432{\pm}.014$ & $0.257{\pm}.021$ & $0.383{\pm}.015$ & $0.228{\pm}.025$ & $0.084{\pm}.004$ & $0.054{\pm}.004$ & $0.095{\pm}.007$ & $0.048{\pm}.013$ \\
LSTM     & $0.418{\pm}.020$ & $0.277{\pm}.024$ & $0.358{\pm}.014$ & $0.204{\pm}.015$ & $0.075{\pm}.004$ & $0.044{\pm}.008$ & $0.088{\pm}.006$ & $0.050{\pm}.010$ \\
PatchTST & $0.425{\pm}.052$ & $0.320{\pm}.060$ & $0.388{\pm}.030$ & $0.281{\pm}.045$ & $0.127{\pm}.022$ & $0.111{\pm}.026$ & $0.139{\pm}.026$ & $0.118{\pm}.030$ \\
EINN     & $0.409{\pm}.010$ & $0.241{\pm}.009$ & $0.364{\pm}.015$ & $0.191{\pm}.025$ & $0.072{\pm}.010$ & $0.039{\pm}.016$ & $0.091{\pm}.009$ & $0.053{\pm}.012$ \\
RNP      & $0.446{\pm}.054$ & $0.353{\pm}.060$ & $0.337{\pm}.063$ & $0.204{\pm}.088$ & $0.094{\pm}.021$ & $0.066{\pm}.019$ & $0.098{\pm}.024$ & $0.0594{\pm}.0234$ \\
EpiFNP   & $0.749{\pm}.075$ & $0.608{\pm}.116$ & $0.619{\pm}.080$ & $0.473{\pm}.106$ & $0.095{\pm}.005$ & $0.062{\pm}.005$ & $0.093{\pm}.003$ & $\mathbf{0.0463}{\pm}.0010$ \\
DSSM     & $0.722{\pm}.058$ & $0.520{\pm}.048$ & $0.528{\pm}.121$ & $0.364{\pm}.123$ & $0.070{\pm}.002$ & $0.032{\pm}.002$ & $0.223{\pm}.054$ & $0.1544{\pm}.0479$ \\
EPI-SST  & $\mathbf{0.345}{\pm}.030$ & $\mathbf{0.220}{\pm}.033$ & $\mathbf{0.334}{\pm}.010$ & $\mathbf{0.182}{\pm}.007$ & $\mathbf{0.065}{\pm}.002$ & $\mathbf{0.028}{\pm}.007$ & $\mathbf{0.087}{\pm}.004$ & $0.0464{\pm}.0080$ \\
\bottomrule
\end{tabular}
\endgroup
\caption{Point accuracy pooled over planning-relevant horizons $h{=}2{:}8$. Mean $\pm$ sample standard deviation over five seeds; lower is better, and the best unrounded mean is bold.}
\label{tab:source_horizon_results}
\end{table*}

\subsection{Structured Horizon Coupling: the Evidence-Driven Renewal Transition}
\label{sec:renewal}

Secondary infections occur over multiple lags and enter surveillance after an additional reporting delay. The renewal equation represents this process through a time-varying reproduction number, a lag-weighted incidence history, and a reporting-delay profile. Renewal-based estimators do not directly provide future multipliers, while latent state-space models such as DSSM~\citep{rangapuram2018deepssm}, ProTran~\citep{tang2021protran}, and USST~\citep{bao2023usst} use generic transitions rather than an epidemic renewal operator. EPI-SST supplies the recursion with horizon-conditioned neural controls and repeatedly applies it to generate the future mean trajectory. The resulting transition retains renewal structure while learning how its internal multiplier and bounded corrections evolve at each forecast horizon. Neural evidence specifies how the transition changes from one forecast step to the next; the trajectory itself remains organized through the multiplier, incidence memory, and delayed observation.

\paragraph{Horizon-conditioned evidence.}
A PatchTST backbone~\citep{nie2022patchtst} splits each channel into
overlapping patches and applies channel-independent self-attention; channel
averaging, positional encoding, and projection produce encoder memory
$\mathbf{E}\in\mathbb{R}^{P\times d_w}$. From a learned $\mathbf{w}_0$, one
evidence state is emitted per step. With
$\mathbf{W}_{<h}=[\mathbf{w}_0,\ldots,\mathbf{w}_{h-1}]$, self-attention
returns
$\mathbf{a}_h=\operatorname{Attn}_{\mathrm{self}}(\mathbf{w}_{h-1},
\mathbf{W}_{<h},\mathbf{W}_{<h})$, cross-attention retrieves
$\mathbf{c}_h=\operatorname{Attn}_{\mathrm{cross}}
(G(\mathbf{w}_{h-1},\mathbf{a}_h),\mathbf{E},\mathbf{E})$, and a fusion block
gives
$\mathbf{w}_h=F_{\theta}(\mathbf{w}_{h-1},\mathbf{a}_h,\mathbf{c}_h,
\operatorname{PE}(h))$.

\paragraph{Renewal transition.}
Fixed normalized lag kernels $g_{1:K}$ and $q_{0:K_d-1}$ carry the two
profiles above, with $K=K_d=6$. The evidence advances a multiplier $R_h$,
latent incidence $i_h$, and delayed signal $o_h$ to the mean:
\begin{equation}
\begin{aligned}
\log R_h
&=\operatorname{clip}\!\left(
\log R_{h-1}+\Delta_h,\;
\log R_{\min},\;\log R_{\max}\right),\\
\log i_h
&=\log R_h
+\operatorname{LSE}_{k=1}^{K}\!\left(\log g_k+\log i_{h-k}\right)
+\varepsilon_h,\\
\log o_h
&=\operatorname{LSE}_{k=0}^{K_d-1}\!\left(\log q_k+\log i_{h-k}\right),\\
\mu_h
&=\alpha\exp(\log o_h)+b+\eta_h,
\quad \alpha=\operatorname{Softplus}(a)>0 .
\end{aligned}
\label{eq:renewal_mean}
\end{equation}
A pooled encoder representation initializes the incidence window, while
$\mathbf{w}_0$ initializes an unconstrained $\log R_0$ that is clipped when
$R_1$ is produced. Running in log space avoids repeated multiplication across
the rollout and keeps latent incidence positive. The mean $\mu_h$ remains
unconstrained because $b$ and $\eta_h$ act after the positive map.

Equation~\ref{eq:renewal_mean} is what couples the horizons. Increments are
applied to the previous value rather than predicted independently, placing
all steps on one autoregressive path, while clipping guarantees
$R_h\in[R_{\min},R_{\max}]$ for every $h\geq1$. Because each update requires
only the current multiplier and six most recent incidence values, the rollout
is a finite-memory recursion in
$\mathbf{s}_h=(\log R_h,\log i_h,\ldots,\log i_{h-5})$ once the evidence is
given. Each coordinate corresponds to an explicitly defined transition
variable, enabling the frozen-path analysis to modify $R_{1:H}$ directly at
a fixed checkpoint. The variables $R_h$, $i_h$, and $o_h$ form the internal
renewal state that generates the forecast mean.

\begin{table*}[!t]
\centering
\begingroup
\tabfont
\setlength{\tabcolsep}{2.5pt}
\begin{tabular}{@{}lccccccccc@{}}
\toprule
& \multicolumn{3}{c}{$h{=}4$} & \multicolumn{3}{c}{$h{=}5$} & \multicolumn{3}{c}{$h{=}6$} \\
\cmidrule(lr){2-4} \cmidrule(lr){5-7} \cmidrule(lr){8-10}
Model & CRPS & CS & LS & CRPS & CS & LS & CRPS & CS & LS \\
\midrule
\multicolumn{10}{@{}l@{}}{\textit{COVID Hospital Admissions}} \\
RNP & $0.233${$\pm$.045} & $0.144${$\pm$.088} & $1.005${$\pm$.196} & $0.248${$\pm$.043} & $0.157${$\pm$.082} & $1.079${$\pm$.181} & $0.261${$\pm$.039} & $0.166${$\pm$.071} & $1.136${$\pm$.160} \\
EpiFNP & $0.360${$\pm$.199} & $0.156${$\pm$.065} & $1.357${$\pm$.615} & $0.510${$\pm$.003} & $0.112${$\pm$.001} & $1.815${$\pm$.007} & $0.512${$\pm$.004} & $0.113${$\pm$.002} & $1.817${$\pm$.008} \\
DSSM & $0.349${$\pm$.023} & $0.078${$\pm$.033} & $1.590${$\pm$.087} & $0.390${$\pm$.031} & $0.097${$\pm$.040} & $1.775${$\pm$.133} & $0.425${$\pm$.043} & $0.103${$\pm$.044} & $1.927${$\pm$.173} \\
EPI-SST & $\mathbf{0.132}${$\pm$.032} & $0.106${$\pm$.030} & $\mathbf{0.533}${$\pm$.147} & $\mathbf{0.150}${$\pm$.023} & $\mathbf{0.095}${$\pm$.039} & $\mathbf{0.650}${$\pm$.109} & $\mathbf{0.177}${$\pm$.015} & $\mathbf{0.057}${$\pm$.022} & $\mathbf{0.794}${$\pm$.075} \\
\midrule
\multicolumn{10}{@{}l@{}}{\textit{COVID Deaths}} \\
RNP & 0.143{$\pm$.063} & 0.136{$\pm$.088} & 0.574{$\pm$.290} & 0.153{$\pm$.069} & 0.141{$\pm$.096} & 0.619{$\pm$.310} & 0.158{$\pm$.066} & 0.142{$\pm$.097} & $\mathbf{0.648}${$\pm$.292} \\
EpiFNP & $0.278${$\pm$.157} & $0.163${$\pm$.032} & $1.159${$\pm$.546} & $0.344${$\pm$.138} & $0.151${$\pm$.031} & $1.421${$\pm$.433} & $0.403${$\pm$.096} & $0.138${$\pm$.018} & $1.624${$\pm$.255} \\
DSSM & $0.223${$\pm$.050} & 0.083{$\pm$.038} & $1.061${$\pm$.174} & $0.258${$\pm$.074} & $\mathbf{0.104}${$\pm$.069} & $1.201${$\pm$.244} & $0.296${$\pm$.101} & $\mathbf{0.125}${$\pm$.099} & $1.352${$\pm$.331} \\
EPI-SST & $\mathbf{0.131}${$\pm$.009} & $\mathbf{0.055}${$\pm$.008} & $\mathbf{0.531}${$\pm$.069} & $\mathbf{0.135}${$\pm$.007} & 0.112{$\pm$.023} & $\mathbf{0.578}${$\pm$.073} & $\mathbf{0.145}${$\pm$.005} & 0.164{$\pm$.032} & 0.665{$\pm$.076} \\
\midrule
\multicolumn{10}{@{}l@{}}{\textit{Measles}} \\
RNP & $0.052${$\pm$.012} & $0.156${$\pm$.062} & $0.098${$\pm$.055} & $0.052${$\pm$.013} & $0.163${$\pm$.059} & $0.104${$\pm$.057} & $0.052${$\pm$.012} & $0.178${$\pm$.060} & $0.107${$\pm$.055} \\
EpiFNP & $0.049${$\pm$.004} & $0.341${$\pm$.045} & $0.142${$\pm$.024} & $0.059${$\pm$.006} & $0.271${$\pm$.018} & $0.200${$\pm$.047} & $0.061${$\pm$.011} & $0.300${$\pm$.045} & $0.225${$\pm$.076} \\
DSSM & $0.026${$\pm$.001} & $0.202${$\pm$.025} & $0.074${$\pm$.005} & $0.028${$\pm$.001} & $0.180${$\pm$.028} & $0.085${$\pm$.005} & $0.030${$\pm$.001} & $0.168${$\pm$.030} & $0.096${$\pm$.005} \\
EPI-SST & $\mathbf{0.019}${$\pm$.005} & $\mathbf{0.141}${$\pm$.052} & $\mathbf{0.036}${$\pm$.003} & $\mathbf{0.020}${$\pm$.004} & $\mathbf{0.130}${$\pm$.067} & $\mathbf{0.039}${$\pm$.004} & $\mathbf{0.023}${$\pm$.005} & $\mathbf{0.126}${$\pm$.042} & $\mathbf{0.044}${$\pm$.004} \\
\midrule
\multicolumn{10}{@{}l@{}}{\textit{Mumps}} \\
RNP & $0.049${$\pm$.017} & $0.198${$\pm$.051} & $0.100${$\pm$.114} & $0.049${$\pm$.017} & $0.197${$\pm$.051} & $0.102${$\pm$.113} & $0.049${$\pm$.017} & $0.197${$\pm$.051} & $0.104${$\pm$.112} \\
EpiFNP & $0.060${$\pm$.002} & $0.367${$\pm$.023} & $0.234${$\pm$.013} & $0.067${$\pm$.002} & $0.331${$\pm$.025} & $0.280${$\pm$.027} & $0.074${$\pm$.003} & $0.358${$\pm$.018} & $0.343${$\pm$.021} \\
DSSM & $0.106${$\pm$.023} & $0.114${$\pm$.022} & $0.454${$\pm$.155} & $0.115${$\pm$.029} & $0.098${$\pm$.029} & $0.496${$\pm$.171} & $0.123${$\pm$.036} & $0.086${$\pm$.036} & $0.542${$\pm$.187} \\
EPI-SST & $\mathbf{0.034}${$\pm$.006} & $\mathbf{0.099}${$\pm$.054} & $\mathbf{0.036}${$\pm$.004} & $\mathbf{0.035}${$\pm$.006} & $\mathbf{0.091}${$\pm$.043} & $\mathbf{0.040}${$\pm$.005} & $\mathbf{0.035}${$\pm$.004} & $\mathbf{0.085}${$\pm$.018} & $\mathbf{0.042}${$\pm$.004} \\
\bottomrule
\end{tabular}
\endgroup
\caption{Horizon-specific probabilistic performance at $h{=}4,5,6$.
Values report the mean $\pm$ sample standard deviation over five seeds;
EPI-SST achieves the lowest mean CRPS for every dataset and reported horizon.}
\label{tab:source_prob_results}
\end{table*}

\paragraph{Bounded neural controls.}
The network corrects the fixed rule at three levels through affine maps of
the evidence: $\Delta_h=c_R\tanh f_R(\mathbf{w}_h)$ controls multiplicative
growth, $\varepsilon_h=c_{\varepsilon}\tanh f_{\varepsilon}(\mathbf{w}_h)$
adjusts incidence beyond the fixed kernel, and
$\eta_h=c_{\eta}\tanh f_{\eta}(\mathbf{w}_h)$ absorbs mismatch between the
delayed signal and reported mean. Separate hyperparameters cap the three
corrections, independently limiting one-step departures in growth, incidence,
and observation mismatch. The controls act at complementary locations:
$\Delta_h$ changes propagation before incidence is generated,
$\varepsilon_h$ adjusts the renewed incidence, and $\eta_h$ adjusts the final
observation mapping. Their separate bounds prevent a single unrestricted
residual from bypassing the renewal state.

\subsection{Uncertainty-Scale Control}
\label{sec:variance_control}
\label{sec:predictive_scale}

Structure on the mean does not by itself give calibrated intervals. Under a
Gaussian likelihood, predicted variance can absorb mean-model error:
inflating around a few large residuals lowers the loss without improving the
forecast. EPI-SST therefore estimates predictive dispersion with a direct
head, leaving the renewal recursion dedicated to mean evolution. The head
maps the same evidence to
$\ell_h=2\log(\operatorname{Softplus}(f_{\sigma}(\mathbf{w}_h))+\delta_0)
+b_{\sigma}$ and $\sigma_h^2=\exp(\ell_h)$, where $f_{\sigma}$ is a two-layer
network, $b_{\sigma}$ is a learned bias, and $\delta_0>0$. Both branches read
$\mathbf{w}_h$ and are trained jointly, while only the renewal branch updates
$\mathbf{s}_h$.

To limit unrestricted variation in predictive dispersion, we shrink
log-variance toward a reference level $m$:
\begin{equation}
\mathcal{L}_{\mathrm{total}}
=
\mathcal{L}_{\mathrm{NLL}}
+\lambda\frac{1}{NH}
\sum_{n=1}^{N}\sum_{h=1}^{H}
\left(\ell_{n,h}-m\right)^2 ,
\label{eq:variance_regularization}
\end{equation}
where $N$ is the number of forecast windows in a minibatch and
$(\lambda,m)$ are selected on validation data only. Training minimizes
Eq.~\ref{eq:variance_regularization} end to end through the unrolled
recursion. The penalty shrinks the overall log-variance level without forcing
it to be constant: deviations from $m$ remain input- and horizon-dependent,
so the emission can represent heteroscedastic uncertainty. Setting
$\lambda=0$ recovers ordinary Gaussian likelihood training, while increasing
$\lambda$ strengthens the shared reference scale. Appendix~A gives the
complete forward pass, numerical log-variance
bounds, and selected values of $(\lambda,m)$.

\section{Experiments}
\label{sec:experiments}

We evaluate EPI-SST on standard forecasting benchmarks, during the abnormal
2019--2020 influenza season, and on whether a trained forecast uses its learned $R_h$
trajectory. The benchmark comparison measures accuracy across continuous and
sparse surveillance targets, the influenza experiment evaluates a sharply
different seasonal trajectory, and the frozen-path interventions test the
mechanism of a fixed trained model. Full dataset statistics, metric definitions, hyperparameter
search, baseline configurations, and exact settings are in Appendix~A.

\subsection{Experimental Setup}

\paragraph{Datasets and baselines.}
We evaluate EPI-SST with context length $L=12$ and forecast horizon $H=8$
on four EpiLearn benchmarks~\citep{liu2024epilearn}: COVID hospital
admissions~\citep{who_dashboard}, COVID deaths~\citep{dong2020interactive},
and the Measles and Mumps records from Project
Tycho~\citep{vanpanhuis2013contagious}. We reserve the 2019--2020 Centers for
Disease Control and Prevention Weighted Influenza-Like Illness (WILI)
season~\citep{cdc_ilinet} for abnormal-regime evaluation. Point baselines
include a convolutional neural network (CNN)~\citep{liu2024epilearn}, long
short-term memory (LSTM) and gated recurrent unit (GRU)
networks~\citep{shahid2020predictions},
DLinear~\citep{zeng2023transformers}, PatchTST~\citep{nie2022patchtst}, and
epidemiologically-informed neural networks
(EINNs)~\citep{rodriguez2022einn}. Probabilistic baselines include the
recurrent attentive neural process (RNP)~\citep{qin2019rnp}, epidemic
functional neural process (EpiFNP)~\citep{kamarthi2021epifnp}, and deep
state-space model (DSSM/DeepState)~\citep{rangapuram2018deepssm}.

\paragraph{Evaluation.}
All models use the same chronological splits, train-derived normalization,
forecast origins, and five seeds. We report root mean squared error (RMSE)
and mean absolute error (MAE) pooled over $h=2{:}8$, and continuous ranked
probability score (CRPS), calibration score (CS), and log score (LS)
separately at $h=4,5,6$. Pooling point errors over weeks 2--8 summarizes the
multiweek planning trajectory beyond the immediate next observation. We keep
probabilistic metrics horizon-specific because coverage and predictive spread
change with lead time; pooling them can conceal over-coverage at one horizon
and under-coverage at another. Complete dataset statistics, metric definitions,
model configurations, and WILI protocols are provided in Appendix~A.

\begin{figure}[!t]
\centering
\includegraphics[width=\columnwidth]{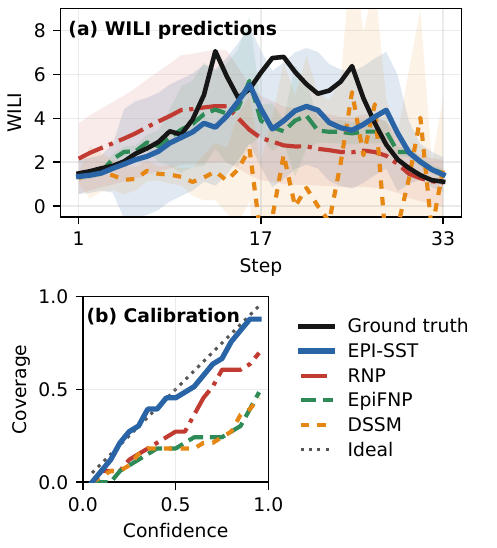}
\caption{Forecasting through the 2019--2020 WILI season using seed 42.
(a) Predictive means and 95\% intervals for 33 three-week-ahead forecasts.
(b) Empirical coverage across nominal confidence levels; the dotted diagonal
denotes ideal calibration.}
\label{fig:wili_anomaly}
\end{figure}

\section{Results}

\subsection{Forecasting Performance Across Standard Regimes}
\label{sec:benchmark_context}

We first compare EPI-SST with six point-forecasting baselines and three
probabilistic baselines using identical data splits, normalization,
forecast origins, and seed indices. The four benchmarks include both
continuous and sparse surveillance targets.

\subsubsection{Point Forecasting}

Point forecasts determine the expected level and timing of epidemic demand,
the quantities most directly used for staffing and capacity planning.
Table~\ref{tab:source_horizon_results} presents pooled point forecasting
performance over the planning-relevant horizons $h{=}2{:}8$. EPI-SST achieves
the lowest RMSE on all four benchmarks, with particularly strong gains on
hospital admissions and Measles: relative to the strongest baseline in each
comparison, it reduces RMSE by $14.2\%$ and $7.1\%$, and MAE by $8.7\%$ and
$12.5\%$, respectively. Hospital admissions and Measles occupy different target
regimes---continuous demand and sparse counts---yet EPI-SST leads both. Its
medium-range accuracy is therefore not confined to one type of surveillance
outcome.

\subsubsection{Probabilistic Forecasting}

Probabilistic forecasts complement this trajectory with uncertainty estimates
needed to assess the risk of exceeding available capacity.
Table~\ref{tab:source_prob_results} presents probabilistic forecasting
performance at $h{=}4,5,6$. EPI-SST achieves the lowest CRPS on all four
datasets at every reported horizon, reducing CRPS by $8\%$--$43\%$ relative
to the strongest baseline for the corresponding dataset and horizon. On
hospital admissions, it also achieves the lowest LS at each horizon and the
lowest CS at $h{=}5,6$; on COVID deaths, it achieves the lowest LS at
$h{=}4,5$. On Measles and Mumps, EPI-SST achieves the best CRPS, CS, and LS
throughout the reported range. The sparse-target CRPS gains therefore
coincide with better calibration and log score.

The advantage extends beyond the three displayed lead times. When all eight
forecast horizons are pooled within each seed, EPI-SST retains the lowest
CRPS and CS on all four datasets and the lowest LS on three. The
horizon-resolved profiles in Appendix~A preserve the medium-range ordering
across the full rollout. Paired hierarchical moving-block bootstrap comparisons
test the same ordering at the forecast-case level: the simultaneous 95\%
intervals for the CRPS improvement over the strongest displayed baseline
exclude zero for every dataset at $h{=}4,5,6$.

\subsection{Forecasting Through an Abnormal Influenza Season}
\label{sec:anomaly}

The 2019--2020 WILI season differs sharply from the seasonal patterns observed
in the training data. We evaluate three-week-ahead forecasts over 33 weeks,
from week 40 of 2019 through week 20 of 2020, including the disruption at the
onset of COVID-19. EPI-SST achieves the lowest RMSE, CRPS, and CS among all
compared models, with values of $1.417$, $0.750$, and $0.037$, respectively.
EpiFNP provides the strongest baseline RMSE and CRPS, while RNP provides the
strongest baseline CS. Relative to these baselines, EPI-SST reduces RMSE by
$2.8\%$, CRPS by $16.9\%$, and calibration error by $78.9\%$.
Across matched seeds 42--44, EPI-SST retains the lowest mean CRPS and CS,
reducing them by $12.0\%$ and $56.7\%$ relative to the corresponding strongest
baselines while matching the best point errors to within $0.5\%$. The
three-seed means preserve the distributional lead shown for seed 42 in
Figure~\ref{fig:wili_anomaly}.

Figure~\ref{fig:wili_anomaly}(a) shows that EPI-SST follows the seasonal rise
and subsequent decline while maintaining predictive intervals through the
volatile portion of the season. Figure~\ref{fig:wili_anomaly}(b) shows that
its coverage curve remains substantially closer to ideal calibration across
most confidence levels, while the baselines consistently under-cover. The
larger reductions in CRPS and CS show that EPI-SST's strongest advantage
during this season lies in the quality of its predictive distribution.

\begin{figure*}[t]
\centering
\includegraphics[width=\textwidth]{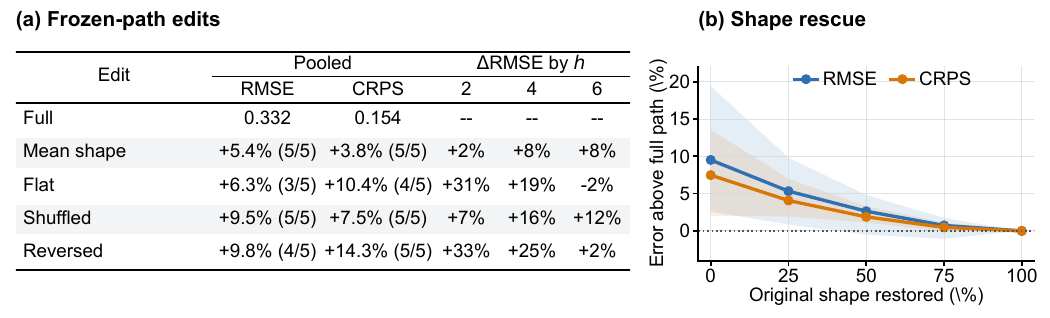}
\caption{Effects of modifying the learned $R_h$ path on COVID hospital
admissions. Level-preserving modifications reduce predictive accuracy (left).
Starting from paths exchanged across forecast windows, progressively restoring
each sample's original shape recovers pooled RMSE and CRPS (right); shaded bands
span the five seed-specific paired changes.}
\label{fig:attribution_audit}
\vspace{0.3em}
\centering
\tabfont
\setlength{\tabcolsep}{5pt}
\begin{tabular}{@{}lccccc@{}}
\toprule
Variant & RMSE & MAE & CRPS & CS & LS \\
\midrule
Full model & $\mathbf{0.332}${$\pm$0.035} & $\mathbf{0.212}${$\pm$0.037} & $\mathbf{0.154}${$\pm$0.027} & $\mathbf{0.042}${$\pm$0.027} & $\mathbf{0.631}${$\pm$0.118} \\
\quad w/o attention conditioning & $0.426${$\pm$0.053} & $0.322${$\pm$0.062} & $0.225${$\pm$0.039} & $0.063${$\pm$0.009} & $1.014${$\pm$0.146} \\
\quad w/o renewal transition (free neural rollout) & $0.437${$\pm$0.184} & $0.319${$\pm$0.208} & $0.225${$\pm$0.132} & $0.101${$\pm$0.033} & $0.930${$\pm$0.501} \\
\quad w/o autoregressive coupling (independent $R_h$) & $0.401${$\pm$0.108} & $0.287${$\pm$0.126} & $0.204${$\pm$0.084} & $0.074${$\pm$0.050} & $0.849${$\pm$0.381} \\
\quad w/o uncertainty-scale control ($\lambda{=}0$) & $0.478${$\pm$0.143} & $0.372${$\pm$0.161} & $0.258${$\pm$0.101} & $0.071${$\pm$0.038} & $1.097${$\pm$0.405} \\
\bottomrule
\end{tabular}
\captionof{table}{Component ablation on COVID hospital admissions, with
metrics pooled over all forecast horizons. Values report the mean $\pm$
sample standard deviation over five seeds; the full model achieves the lowest
mean on all five metrics.}
\label{tab:ablation}
\end{figure*}

\subsection{Analysis of the Learned Renewal Path}
\label{sec:attribution}

\paragraph{Path dependence.}
To examine whether EPI-SST uses the learned renewal path, we modify only the
predicted $R_{1:H}$ trajectory and recompute the forecast while holding all
model parameters and other states fixed. The original trajectory reproduces
the stored predictions to within $3\times10^{-6}$. Each modification
preserves the geometric mean of $R$ within a forecast window, isolating
temporal shape from overall level. We replace the original path with the mean
path computed from validation data, exchange paths between forecast windows,
reverse their temporal order, or flatten them. We also progressively restore
the original path to measure how forecast accuracy changes with its learned
shape.

Because the model is not retrained, changes in accuracy directly measure its
dependence on the learned path.

Replacing the original path with the mean path computed from validation data
increases pooled RMSE by $5.4\%$ for all five seeds. Exchanging paths between
forecast windows increases RMSE by $9.5\%$ and CRPS by $7.5\%$, again for all
five seeds (Figure~\ref{fig:attribution_audit}(a)). Reversing each path
preserves its level, variance, and set of values, but increases CRPS by
$14.3\%$ for all five seeds and RMSE by $9.8\%$ for four of five seeds.
Flattening or reversing the paths increases RMSE by $31$--$33\%$ at $h{=}2$
and by $19$--$25\%$ at $h{=}4$. Starting from exchanged paths, restoring
$25\%$, $50\%$, $75\%$, and $100\%$ of each sample's original shape reduces
excess pooled RMSE from $9.5\%$ to $5.3\%$, $2.6\%$, $0.7\%$, and zero;
CRPS follows the same monotone recovery, from $7.5\%$ to $4.1\%$, $1.9\%$,
$0.5\%$, and zero (Figure~\ref{fig:attribution_audit}(b)). Forecasts therefore
depend on the sample-specific shape and temporal order of the learned path.

\paragraph{Directional response.}
To examine the direction of this dependence, we increase $R_h$ either at the
first forecast step or across the full path. A $20\%$ increase at $R_1$
produces a mean forecast change of $0.016$ at $h{=}1$, reaches $0.027$ at
$h{=}3$, and remains nonzero at $h{=}8$. Applying the same increase across
the full path produces changes of $0.016$, $0.066$, $0.074$, and $0.072$ at
$h{=}1,4,6,8$, respectively. These values are averaged over evaluation
windows and then across five seeds. The response is monotone across
intervention magnitudes and consistent across all five seeds; decreasing
$R_h$ reverses its direction. Combined with the path-restoration experiment,
the signed response identifies a consistent and measurable effect of the
learned $R_h$ trajectory on the predictive mean.

\subsection{Component Ablation}
\label{sec:ablation}

Table~\ref{tab:ablation} evaluates four central design choices on COVID
hospital admissions. Every ablation uses the same data, forecast origins, and
seed indices as the full model.
The full model achieves the lowest five-seed mean on all five metrics.
Relative to the full model, removing attention conditioning---and therefore
encoder-memory retrieval from the evidence rollout---increases RMSE by
$28.4\%$ and CRPS by $45.8\%$. Replacing the renewal transition with a free
neural rollout increases RMSE by $31.6\%$ and CRPS by $46.1\%$. Predicting
independent $R_h$ values instead of an autoregressively coupled path increases
RMSE by $20.8\%$ and CRPS by $32.5\%$, while removing uncertainty-scale
control increases them by $44.0\%$ and $67.5\%$, respectively. The
no-attention variant increases both RMSE and CRPS for four of the five seeds,
despite retaining the PatchTST encoder and renewal transition. Each ablation
also produces a worse five-seed mean than the full model on every reported
metric. Attention conditioning, the renewal transition, autoregressive horizon
coupling, and uncertainty-scale control therefore account for distinct parts
of the full model's performance.

\section{Discussion and Societal Impact}

EPI-SST makes the renewal equation a forward transition within a learned
multi-horizon rollout. The transition generates a coupled mean trajectory,
while the emission head sets predictive uncertainty. Across continuous and
sparse surveillance targets, this division supports accurate mean trajectories
and consistently strong CRPS, including during the abnormal influenza season.

The ablations locate these gains in three interacting choices: renewal
recursion, autoregressive horizon coupling, and attention-conditioned evidence.
Frozen-path edits ask a different question by changing only the trained
$R_{1:H}$ trajectory. Accuracy falls under level-preserving changes to shape
and order, returns as the learned path is restored, and responds monotonically
to signed perturbations. The architecture benefits from renewal recursion and
horizon coupling, and the trained model uses the ordered state produced by
that transition. Removing uncertainty-scale control instead produces the
largest CRPS degradation, matching its distinct role in the emission path.

Together, the two paths explain the complementary gains. The renewal state
couples future means through incidence memory, while the emission path
converts horizon-conditioned evidence into predictive spread, jointly
supporting multiweek accuracy and distributional quality.

\paragraph{Societal impact.}
Probabilistic epidemic forecasts support staffing, inventory, and capacity
decisions that depend on both expected demand and threshold-exceedance risk.
EPI-SST produces both in one rollout and exposes the state path driving its
mean forecast. Reporting delays,
revisions, and changes in care-seeking behavior still require deployment-time
monitoring. The learned $R_h$ trajectory is a forecasting state; causal or
policy conclusions require separate epidemiological identification.

\paragraph{Scope and future directions.}
The abnormal-season analysis covers one influenza season, with three-seed
replication. The main experiments share a fixed lag profile across diseases;
hospital-admission sweeps support its shape, width, bounds, and correction
limits within the tested neighborhood. Disease-specific kernels could reflect
differences between transmission and observation processes while retaining the
renewal-style transition. The current emissions are horizon-specific;
multivariate emissions could model cross-horizon dependence.

\clearpage

\bibliography{refs}
\clearpage
\appendix
\raggedbottom

\section{A. Dataset Construction and Preprocessing}
\label{sec:app_data}

We use four multi-region epidemic-surveillance bundles distributed with
EpiLearn~\citep{liu2024epilearn}: COVID hospital
admissions~\citep{who_dashboard}, COVID deaths~\citep{dong2020interactive},
and the Measles and Mumps records from Project
Tycho~\citep{vanpanhuis2013contagious}. We additionally use CDC Weighted
Influenza-Like Illness (WILI)~\citep{cdc_ilinet} for the abnormal-season
analysis. Table~\ref{tab:dataset_details} gives the retained region counts
and the number of forecast windows generated from each chronological
partition. The standard benchmarks use the provided 70\%/10\%/20\%
train/validation/test partition; the 2019--2020 WILI season is reserved as
the abnormal-regime test period. Region-wise normalization statistics are
estimated from the training segment and applied unchanged to validation and
test data. The Tycho experiments retain the regions marked active by the
provided preprocessing metadata.

\begin{center}
\centering
\tabfont
\setlength{\tabcolsep}{2.5pt}
\begin{tabular}{@{}lcrrr@{}}
\toprule
Dataset & Regions & Train & Validation & Test \\
\midrule
COVID Hospital Adm. & 49 & 7{,}099 & 260 & 1{,}409 \\
COVID Deaths & 105 & 14{,}650 & 468 & 2{,}838 \\
Measles & 51 & 129{,}516 & 17{,}666 & 36{,}349 \\
Mumps & 51 & 59{,}728 & 7{,}699 & 16{,}404 \\
WILI (2019--20 holdout) & 1 & 683 & 33 & 33 \\
\bottomrule
\end{tabular}
\captionof{table}{Dataset statistics and chronological forecast-window counts.}
\label{tab:dataset_details}
\end{center}

\section{B. Evaluation Protocol and Metrics}
\label{sec:app_evaluation}

All standard-regime comparisons use context length $L=12$, maximum forecast
horizon $H=8$, and seeds $\{42,\ldots,46\}$. Models evaluated on a given
dataset share the chronological split, train-derived normalization, window
stride, test forecast origins, and seed indices. The WILI figure in the main
text displays seed 42 over 33 common rolling origins; the replication in
Table~\ref{tab:wili_multiseed} evaluates the same origins for seeds 42--44.
Reported uncertainty is the sample standard deviation across seeds and
therefore describes training stability.

CNN, DLinear, GRU, LSTM, PatchTST, and EINN use their EpiLearn
implementations; RNP and EpiFNP use the released reference code, and DSSM
uses GluonTS DeepState.

\begin{figure*}[t]
\centering
\includegraphics[width=0.90\textwidth]{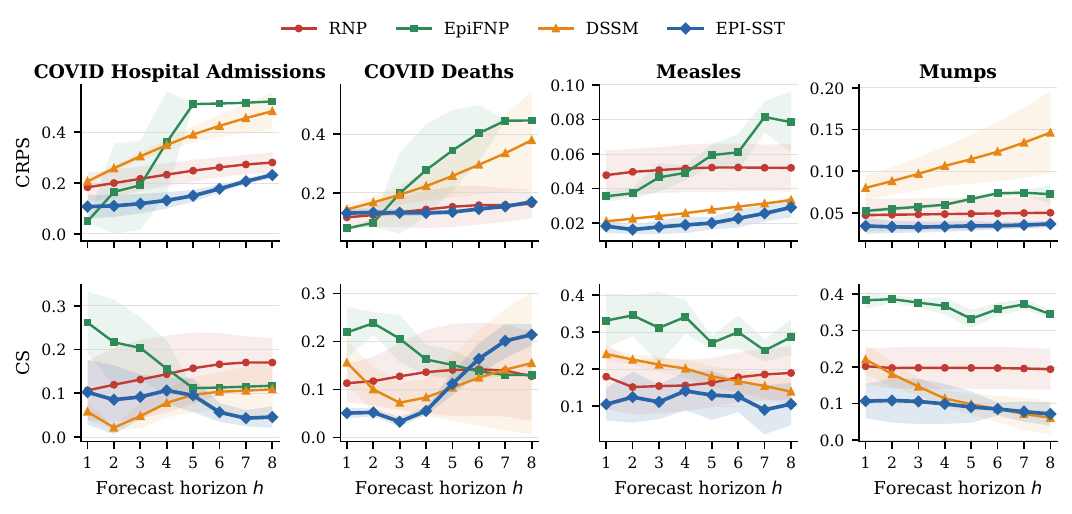}
\caption{Horizon-resolved CRPS and CS over $h=1{:}8$ on four benchmarks.
Lines are five-seed means and bands show $\pm1$ sample standard deviation;
$h=4,5,6$ are reported explicitly in the main text.}
\label{fig:full_horizon_prob}
\end{figure*}

For targets $y_i$, predictive means $\hat\mu_i$, predictive CDFs $F_i$, and
LS half-width $\delta=0.3$, the four standard-benchmark metrics are computed
in normalized target space and lower is better. WILI is evaluated separately
on its raw scale. Because WILI is nonnegative, all displayed models are
evaluated under the same observation transform $Y=\max(0,Z)$: point means,
draws, and quantiles are projected onto nonnegative support, corresponding to
a Gaussian distribution left-censored at zero. We use
\begin{equation}
\begin{aligned}
\mathrm{RMSE}&=\sqrt{N^{-1}\sum_{i=1}^{N}(y_i-\hat\mu_i)^2},\\[-2pt]
\mathrm{MAE}&=N^{-1}\sum_{i=1}^{N}|y_i-\hat\mu_i|,\\[-2pt]
\mathrm{CRPS}&=N^{-1}\sum_{i=1}^{N}\int_{-\infty}^{\infty}
\bigl(F_i(x)-\mathbf{1}\{x\ge y_i\}\bigr)^2\,dx,\\[-2pt]
\mathrm{LS}&=-N^{-1}\sum_{i=1}^{N}\log
\Pr_{F_i}(|Y-y_i|\le\delta).
\end{aligned}
\label{eq:app_metrics}
\end{equation}
For a nominal central interval with coverage $c$, let $k_c$ be its empirical
coverage. Following EpiFNP~\citep{kamarthi2021epifnp}, the calibration score
is
\begin{equation}
\mathrm{CS}=\frac{1}{99}\sum_{c\in\{0.01,\ldots,0.99\}}|k_c-c|.
\label{eq:app_cs}
\end{equation}
CRPS is evaluated in closed form for Gaussian predictions. For
horizon-specific results, regions and forecast windows are pooled at a fixed
$h$ within each seed before summarizing across seeds. For pooled results, the
stated horizons are concatenated within each seed before the metric is
computed. The main probabilistic table keeps $h=4,5,6$ separate because
coverage errors at different lead times can cancel when aggregated;
Table~\ref{tab:global_pooled_prob} and Figure~\ref{fig:full_horizon_prob}
provide complementary pooled and lead-resolved views.

For the main-table CRPS comparisons, we additionally average case-level CRPS
over the five matched seeds and apply a paired hierarchical moving-block
bootstrap with 10{,}000 replicates. Regions are resampled first, followed by
circular blocks of four consecutive forecast origins within each selected
region; identical draws are used for EPI-SST and its comparator. We compare
EPI-SST with the lowest-mean displayed baseline in each dataset--horizon cell
and use Bonferroni simultaneous 95\% intervals across the 12 comparisons.

\begin{table*}[t]
\centering
\begingroup
\tabfont
\setlength{\tabcolsep}{3.2pt}
\begin{tabular}{@{}lcccccc@{}}
\toprule
& \multicolumn{3}{c}{Hospital Admissions} & \multicolumn{3}{c}{Deaths} \\
\cmidrule(lr){2-4}\cmidrule(lr){5-7}
Model & CRPS & CS & LS & CRPS & CS & LS \\
\midrule
RNP & 0.237{$\pm$.044} & 0.140{$\pm$.075} & 1.022{$\pm$.187} & 0.144{$\pm$.052} & 0.125{$\pm$.075} & $\mathbf{0.583}${$\pm$.235} \\
EpiFNP & 0.353{$\pm$.060} & 0.075{$\pm$.023} & 2.941{$\pm$.232} & 0.287{$\pm$.053} & 0.114{$\pm$.023} & 2.829{$\pm$.183} \\
DSSM & 0.359{$\pm$.031} & 0.061{$\pm$.037} & 1.626{$\pm$.117} & 0.249{$\pm$.072} & 0.087{$\pm$.045} & 1.158{$\pm$.246} \\
EPI-SST & $\mathbf{0.154}${$\pm$.027} & $\mathbf{0.042}${$\pm$.027} & $\mathbf{0.631}${$\pm$.118} & $\mathbf{0.141}${$\pm$.008} & $\mathbf{0.081}${$\pm$.014} & 0.615{$\pm$.070} \\
\midrule
& \multicolumn{3}{c}{Measles} & \multicolumn{3}{c}{Mumps} \\
\cmidrule(lr){2-4}\cmidrule(lr){5-7}
Model & CRPS & CS & LS & CRPS & CS & LS \\
\midrule
RNP & 0.051{$\pm$.013} & 0.167{$\pm$.071} & 0.097{$\pm$.059} & 0.049{$\pm$.019} & 0.197{$\pm$.056} & 0.101{$\pm$.128} \\
EpiFNP & 0.056{$\pm$.002} & 0.288{$\pm$.012} & 1.593{$\pm$.038} & 0.064{$\pm$.001} & 0.361{$\pm$.004} & 1.744{$\pm$.027} \\
DSSM & 0.027{$\pm$.001} & 0.190{$\pm$.024} & 0.079{$\pm$.005} & 0.111{$\pm$.029} & 0.118{$\pm$.025} & 0.481{$\pm$.168} \\
EPI-SST & $\mathbf{0.021}${$\pm$.004} & $\mathbf{0.086}${$\pm$.061} & $\mathbf{0.037}${$\pm$.004} & $\mathbf{0.035}${$\pm$.006} & $\mathbf{0.082}${$\pm$.043} & $\mathbf{0.038}${$\pm$.004} \\
\bottomrule
\end{tabular}
\endgroup
\caption{Pooled probabilistic metrics over all test regions and $h=1{:}8$ in normalized target space. Pooled CS complements rather than replaces lead-specific calibration. Entries are mean $\pm$ sample standard deviation over five seeds; the best mean for each dataset and metric is bold.}
\label{tab:global_pooled_prob}
\end{table*}

\section{C. EPI-SST Configuration, Training, and Computational Resources}
\label{sec:app_configuration}

For the four standard benchmarks, EPI-SST uses a PatchTST encoder with
$d_{\mathrm{model}}=64$, four attention
heads, two Transformer layers, patch length 4, stride 2, feed-forward
multiplier 4, and dropout 0.1. The evidence-rollout width is
$d_w=d_{\mathrm{model}}/2=32$. The same model is shared across regions within
each dataset, and each forecast is conditioned only on its normalized
surveillance history. At each patch position,
the encoder adds stride-indexed positional encoding, applies
layer normalization, and projects the target representation to the evidence
width. The pooled representation initializes the
$K$ log-incidence values preceding the forecast origin, while the learned
evidence state initializes $\log R_0$.

Algorithm~\ref{alg:rollout} gives the complete future rollout. The encoder is
called once per input window; subsequent steps update horizon-conditioned
evidence, advance the finite-memory renewal state, and emit Gaussian
marginals. Training differentiates through the full unrolled recursion and
uses no latent sampling or variational approximation.

\begin{algorithm}[t]
\caption{EPI-SST forward pass for one input window}
\label{alg:rollout}
\begin{algorithmic}[1]
\REQUIRE history $\mathbf{X}$; kernels $g_{1:K},q_{0:K_d-1}$; caps
$c_R,c_\varepsilon,c_\eta$; bounds $R_{\min},R_{\max},\ell_{\min},\ell_{\max}$
\ENSURE $\{(\mu_h,\sigma_h^2)\}_{h=1}^{H}$
\STATE $\mathbf{E}\leftarrow\operatorname{PatchTST}(\mathbf{X})$
\STATE $\bar{\mathbf e}\leftarrow\operatorname{pool}(\mathbf{E})$;
$(\log i_{1-K},\ldots,\log i_0)\leftarrow\operatorname{Init}_i(\bar{\mathbf e})$
\STATE $\mathbf{w}_0\leftarrow$ learned initial evidence state
\STATE $\log R_0\leftarrow f_{R_0}(\mathbf{w}_0)$
\FOR{$h=1$ \TO $H$}
  \STATE $\mathbf{W}_{<h}\leftarrow[\mathbf{w}_0,\ldots,\mathbf{w}_{h-1}]$
  \STATE $\mathbf{a}_h\leftarrow
  \operatorname{Attn}_{\mathrm{self}}(\mathbf{w}_{h-1},\mathbf{W}_{<h},\mathbf{W}_{<h})$
  \STATE $\mathbf{c}_h\leftarrow
  \operatorname{Attn}_{\mathrm{cross}}(G(\mathbf{w}_{h-1},\mathbf{a}_h),\mathbf{E},\mathbf{E})$
  \STATE $\mathbf{w}_h\leftarrow F_\theta(\mathbf{w}_{h-1},\mathbf{a}_h,
  \mathbf{c}_h,\operatorname{PE}(h))$
  \STATE $\Delta_h\leftarrow c_R\tanh f_R(\mathbf{w}_h)$
  \STATE $\log R_h\leftarrow\operatorname{clip}(\log R_{h-1}+\Delta_h,
  \log R_{\min},\log R_{\max})$
  \STATE $\log i_h\leftarrow\log R_h+
  \operatorname{LSE}_{k=1}^{K}(\log g_k+\log i_{h-k})
  +c_\varepsilon\tanh f_\varepsilon(\mathbf{w}_h)$
  \STATE $\log o_h\leftarrow\operatorname{LSE}_{k=0}^{K_d-1}
  (\log q_k+\log i_{h-k})$
  \STATE $\mu_h\leftarrow\alpha\exp(\log o_h)+b+
  c_\eta\tanh f_\eta(\mathbf{w}_h)$
  \STATE $\ell_h\leftarrow2\log
  (\operatorname{Softplus}(f_\sigma(\mathbf{w}_h))+\delta_0)+b_\sigma$
  \STATE $\sigma_h^2\leftarrow
  \exp(\operatorname{clip}(\ell_h,\ell_{\min},\ell_{\max}))$
\ENDFOR
\RETURN $\{\mathcal{N}(\mu_h,\sigma_h^2)\}_{h=1}^{H}$
\end{algorithmic}
\end{algorithm}

The transition uses $K=K_d=6$, $R_h\in[0.1,4]$, $c_R=0.7$, and
$c_\varepsilon=c_\eta=0.3$. The fixed normalized lag kernels are
\begin{equation}
g=q=(0.1859,0.2255,0.2052,0.1659,0.1258,0.0916),
\end{equation}
obtained by normalizing $j\exp(-j/2)$ for $j=1,\ldots,6$. For $g$, entries
refer to lags $1,\ldots,6$; for $q$, they refer to lags $0,\ldots,5$. These
generic kernels are shared unchanged across diseases and are not
disease-specific epidemiological distributions or validation-selected
parameters. Their role is to provide fixed normalized lag structure for the
renewal-style rollout. Kernel width and shape remain fixed in all main-text
comparisons; a controlled study later in this appendix audits changes in
shape while retaining the same width.

Training uses AdamW with batch size 32, zero weight decay, at most 100 epochs,
and early stopping on validation total loss with patience 10. The model has
126{,}414 trainable parameters. Renewal and delay convolutions are evaluated
in log space with log-sum-exp; $10^{-6}$ is added before logarithms in the
scale head, and likelihood evaluation clips log-variance to $[-10,2]$.
Observed emitted log-variances lie within $[-7.7,0.2]$, so the clipping bounds
are inactive for the reported predictions.

The separately selected WILI case-study model uses $L=8$, $H=4$, patch length
3, stride 1, feed-forward multiplier 2, dropout 0.05, and evidence width
$d_w=128$. It has 440{,}462 trainable parameters and is trained for at most 60
epochs with patience 10 and early stopping on validation MSE. Its renewal
kernel, multiplier-bound, and correction-cap settings are unchanged from the
standard configuration above.

Experiments run on one NVIDIA RTX A6000 GPU (48~GB) under Linux 6.8 on
x86-64, Python 3.13.13, PyTorch 2.5.1+cu121, CUDA 12.1, and cuDNN 9.1. A
complete EPI-SST training run takes approximately 8.8 minutes and uses
584~MiB peak GPU memory in the reference configuration.

\section{D. Validation-Based Hyperparameter Selection}
\label{sec:app_selection}

Hyperparameters are selected on validation data using a coarse-to-fine
procedure. We initialize the log-variance reference at
$m_0=\log(\mathrm{MSE}_{\mathrm{val}}+10^{-8})$, screen coarse grids over the
learning-rate scale, $\lambda\in[0,0.5]$, and $m\in[-12,0]$, and then refine
the most promising neighborhoods. Candidate configurations are compared
jointly using validation RMSE, CRPS, NLL, and CS, with point accuracy used to
break close trade-offs. The selected configuration is fixed before the five
reported seed runs. Table~\ref{tab:varreg_hyperparams} lists the final values;
$m$ is the reference log-variance and $\lambda$ weights its regularizer.

\begin{center}
\centering
\tabfont
\setlength{\tabcolsep}{4.5pt}
\begin{tabular}{@{}lcc@{}}
\toprule
Dataset & $\lambda$ & $m$ \\
\midrule
COVID Hospital Admissions & 0.036 & $-11.0$ \\
COVID Deaths & 0.001 & $-0.785$ \\
Measles & 0.5 & $-8.25$ \\
Mumps & 0.5 & $-6.75$ \\
WILI & 0.010 & $-2.0$ \\
\bottomrule
\end{tabular}
\captionof{table}{Validation-selected EPI-SST uncertainty-scale settings.}
\label{tab:varreg_hyperparams}
\end{center}

To isolate sensitivity to the scale penalty, we vary $\lambda$ on the
hospital-admission validation set while fixing $m=-11$ and all remaining
settings. Table~\ref{tab:lambda_sensitivity} reports the three-seed means.
From no scale penalty through $\lambda=0.12$, CRPS remains between $0.114$
and $0.118$, while CS and LS expose the calibration--sharpness trade-off. The
selected $\lambda=0.036$ attains the lowest LS and ties the lowest CRPS to the
shown precision without being chosen on test data.

\begin{center}
\centering
\tabfont
\setlength{\tabcolsep}{4.2pt}
\begin{tabular}{@{}lccc@{}}
\toprule
$\lambda$ & CRPS & CS & LS \\
\midrule
0.000 & 0.118 & 0.121 & 0.451 \\
0.010 & 0.115 & 0.118 & 0.437 \\
0.020 & 0.118 & 0.079 & 0.428 \\
0.036$^\dagger$ & $\mathbf{0.114}$ & 0.056 & $\mathbf{0.418}$ \\
0.060 & $\mathbf{0.114}$ & 0.062 & 0.457 \\
0.120 & 0.116 & $\mathbf{0.045}$ & 0.548 \\
\addlinespace[0.25em]
\multicolumn{4}{@{}l@{}}{\small $^\dagger$ Selected setting.} \\
\bottomrule
\end{tabular}
\captionof{table}{Validation sensitivity to uncertainty-scale regularization
on hospital admissions, averaged over seeds 42--44 with $m=-11$.}
\label{tab:lambda_sensitivity}
\end{center}

The remaining analyses address four complementary questions behind the main
claims. The extended probabilistic results test whether the reported
medium-range gains continue across the full rollout and under pooled
aggregation. The WILI replication tests whether the abnormal-season ordering
persists across training seeds. The protocol sweeps test whether performance
is tied to a single context length or requested horizon, while the structural
sweeps isolate the effects of lag design, multiplier bounds, and bounded
corrections.

\section{E. Extended Probabilistic Results}
\label{sec:app_pooled_prob}

Table~\ref{tab:global_pooled_prob} pools probabilistic predictions over
$h=1{:}8$ within each seed. This aggregate view complements the
horizon-specific $h=4,5,6$ results in the main text: pooled CS summarizes
overall calibration, whereas the main table preserves lead-specific
over- and under-coverage that aggregation can conceal.

EPI-SST leads 11 of the 12 pooled dataset--metric comparisons: it has the
lowest pooled CRPS and CS on all four datasets and the lowest LS on three,
with RNP leading only deaths LS. On the sparse Measles and Mumps targets,
EPI-SST leads all six pooled CRPS, CS, and LS comparisons.

Figure~\ref{fig:full_horizon_prob} complements the pooled summary by resolving
CRPS and calibration separately at every lead time. It confirms that the
medium-range entries in the main text are part of continuous horizon profiles
rather than artifacts of pooling across lead times.

Table~\ref{tab:main_crps_bootstrap} reports paired CRPS differences for the
main table. Positive values favor EPI-SST. All 12 simultaneous intervals
exclude zero, including the smaller but consistently positive improvements on
COVID deaths.

\begin{center}
\centering
\tabfont
\setlength{\tabcolsep}{0.35pt}
\begin{tabular}{@{}lccc@{}}
\toprule
Dataset & $h{=}4$ & $h{=}5$ & $h{=}6$ \\
\midrule
Hospital Adm. & \shortstack{$0.101$\\$[0.080,0.122]$} & \shortstack{$0.099$\\$[0.079,0.117]$} & \shortstack{$0.084$\\$[0.067,0.101]$} \\
COVID Deaths & \shortstack{$0.012$\\$[0.001,0.022]$} & \shortstack{$0.018$\\$[0.008,0.027]$} & \shortstack{$0.013$\\$[0.005,0.021]$} \\
Measles & \shortstack{$0.0068$\\$[0.0035,0.0098]$} & \shortstack{$0.0077$\\$[0.0044,0.0108]$} & \shortstack{$0.0068$\\$[0.0038,0.0099]$} \\
Mumps & \shortstack{$0.0149$\\$[0.0115,0.0182]$} & \shortstack{$0.0146$\\$[0.0114,0.0179]$} & \shortstack{$0.0148$\\$[0.0117,0.0181]$} \\
\bottomrule
\end{tabular}
\captionof{table}{Paired CRPS improvement of EPI-SST over the strongest
displayed baseline after averaging case-level losses over five matched runs. Brackets give Bonferroni
simultaneous 95\% bootstrap intervals over all 12 comparisons.}
\label{tab:main_crps_bootstrap}
\end{center}

\section{F. WILI Multi-Seed Replication}
\label{sec:app_wili_multiseed}

The main text uses one seed to visualize the complete 2019--2020 trajectory
and its predictive intervals. To test whether its distributional ordering is
specific to that checkpoint, we repeat the same calendar-matched experiment
for seeds 42--44, keeping the 33 forecast origins identical across methods.

\begin{center}
\centering
\tabfont
\setlength{\tabcolsep}{2.2pt}
\begin{tabular}{@{}lcccc@{}}
\toprule
Model & RMSE & MAE & CRPS & CS \\
\midrule
RNP & $2.008{\pm}.035$ & $1.513{\pm}.034$ & $1.208{\pm}.013$ & $0.134{\pm}.038$ \\
EpiFNP & $\mathbf{1.452}{\pm}.054$ & $\mathbf{1.006}{\pm}.056$ & $0.884{\pm}.041$ & $0.254{\pm}.048$ \\
DSSM & $2.818{\pm}.445$ & $2.268{\pm}.375$ & $1.786{\pm}.304$ & $0.245{\pm}.075$ \\
EPI-SST & $1.459{\pm}.057$ & $1.011{\pm}.031$ & $\mathbf{0.778}{\pm}.060$ & $\mathbf{0.058}{\pm}.019$ \\
\bottomrule
\end{tabular}
\captionof{table}{Abnormal 2019--2020 WILI results over matched seeds
42--44 and identical forecast origins. Entries are mean $\pm$ sample standard
deviation on the raw WILI scale under the common zero-censored observation
model; the best mean per metric is bold.}
\label{tab:wili_multiseed}
\end{center}

EPI-SST retains the lowest mean CRPS and CS across the three runs, reducing
CRPS by $12.0\%$ relative to EpiFNP and CS by $56.7\%$ relative to RNP. Its
mean RMSE and MAE are within $0.5\%$ of EpiFNP. The replication therefore
supports the distributional and calibration advantage highlighted by the
main figure while separating that claim from universal point dominance on
this season.

\section{G. Protocol Sensitivity}
\label{sec:app_sensitivity}

To test dependence on the selected protocol, we vary one axis at a time
around $(L,H)=(12,8)$ on COVID hospital admissions, using matched splits,
forecast origins, and seeds 42--44. The context sweep fixes $H=8$ and tests
$L\in\{8,10,12,14\}$; $L=12$ leads all four metrics. The horizon sweep fixes
$L=12$ and tests $H\in\{4,6,8,10\}$. RMSE, CRPS, and LS increase with the
requested rollout length, while CS is nonmonotonic because it measures
calibration rather than difficulty alone.

\begin{center}
\centering
\tabfont
\setlength{\tabcolsep}{1.2pt}
\begin{tabular}{@{}lcccc@{}}
\toprule
Setting & RMSE & CRPS & CS & LS \\
\midrule
\multicolumn{5}{@{}l}{\textit{Context length ($H=8$)}} \\
$L=8$ & $0.410{\pm}0.107$ & $0.214{\pm}0.084$ & $0.056{\pm}0.071$ & $0.906{\pm}0.370$ \\
$L=10$ & $0.446{\pm}0.125$ & $0.244{\pm}0.095$ & $0.104{\pm}0.067$ & $1.036{\pm}0.416$ \\
$L=12^\dagger$ & $0.318{\pm}0.013$ & $0.144{\pm}0.014$ & $0.029{\pm}0.018$ & $0.585{\pm}0.065$ \\
$L=14$ & $0.367{\pm}0.079$ & $0.165{\pm}0.056$ & $0.107{\pm}0.087$ & $0.662{\pm}0.209$ \\
\midrule
\multicolumn{5}{@{}l}{\textit{Forecast horizon ($L=12$)}} \\
$H=4$ & $0.229{\pm}0.015$ & $0.088{\pm}0.006$ & $0.101{\pm}0.022$ & $0.301{\pm}0.024$ \\
$H=6$ & $0.288{\pm}0.013$ & $0.122{\pm}0.014$ & $0.064{\pm}0.027$ & $0.477{\pm}0.057$ \\
$H=8^\dagger$ & $0.318{\pm}0.013$ & $0.144{\pm}0.014$ & $0.029{\pm}0.018$ & $0.585{\pm}0.065$ \\
$H=10$ & $0.466{\pm}0.143$ & $0.240{\pm}0.108$ & $0.072{\pm}0.084$ & $0.974{\pm}0.427$ \\
\addlinespace[0.25em]
\multicolumn{5}{@{}l@{}}{\small $^\dagger$ Primary setting.} \\
\bottomrule
\end{tabular}
\captionof{table}{Protocol sensitivity on hospital admissions, varying one
axis at a time around $(L,H)=(12,8)$. Entries are mean $\pm$ sample standard
deviation over seeds 42--44.}
\label{tab:hospital_protocol_sensitivity}
\end{center}

\section{H. Structural Sensitivity}
\label{sec:app_structural_sensitivity}

With matched hospital-admission splits, origins, and seeds 42--44, we vary one
structural choice at a time. The lag-shape sweep uses
$g_j(\tau)=q_j(\tau)\propto j\exp(-j/\tau)$ for $\tau\in\{1,2,3\}$ at
$K=K_d=6$. The prespecified $\tau=2$ profile is compared with sharper and
flatter alternatives; no setting is selected using test performance.

\begin{center}
\centering
\tabfont
\setlength{\tabcolsep}{0.8pt}
\begin{tabular}{@{}lccccc@{}}
\toprule
Kernel & RMSE & MAE & CRPS & CS & LS \\
\midrule
$\tau=1$ & \shortstack{$0.396$\\$\pm .132$} & \shortstack{$0.283$\\$\pm .150$} & \shortstack{$0.203$\\$\pm .102$} & \shortstack{$0.089$\\$\pm .075$} & \shortstack{$0.846$\\$\pm .449$} \\
$\tau=2$ & \shortstack{$\mathbf{0.318}$\\$\pm .013$} & \shortstack{$\mathbf{0.199}$\\$\pm .021$} & \shortstack{$\mathbf{0.144}$\\$\pm .014$} & \shortstack{$\mathbf{0.029}$\\$\pm .018$} & \shortstack{$\mathbf{0.585}$\\$\pm .065$} \\
$\tau=3$ & \shortstack{$0.391$\\$\pm .128$} & \shortstack{$0.274$\\$\pm .148$} & \shortstack{$0.198$\\$\pm .102$} & \shortstack{$0.087$\\$\pm .071$} & \shortstack{$0.836$\\$\pm .467$} \\
\bottomrule
\end{tabular}
\captionof{table}{Hospital test metrics pooled over $h=1{:}8$; cells show
mean above $\pm$ sample standard deviation over seeds 42--44.}
\label{tab:kernel_shape_sensitivity}
\end{center}

The $\tau=2$ reruns reproduce the main checkpoints and lead all validation
and test means. Relative to $\tau=1$ and $3$, they reduce test RMSE by 19.6\%
and 18.6\% and CRPS by 29.2\% and 27.3\%; MAE, CS, and LS also lead, with CS
lower in every run.

Matched $h=1{:}8$ sweeps next vary the $j\exp(-j/2)$ kernel width, multiplier
bounds, and correction caps. The primary setting leads mean RMSE and CRPS
throughout; removing either correction worsens both, particularly observation
correction.

\begin{center}
\centering
\tabfont
\renewcommand{\arraystretch}{0.80}
\setlength{\tabcolsep}{1.5pt}
\begin{tabular}{@{}llcc@{}}
\toprule
Factor & Setting & RMSE & CRPS \\
\midrule
\textit{Width} & $K=3$ & $0.384{\pm}.089$ & $0.191{\pm}.074$ \\
& $K=6$ & $\mathbf{0.318}{\pm}.013$ & $\mathbf{0.144}{\pm}.014$ \\
& $K=9$ & $0.381{\pm}.092$ & $0.192{\pm}.071$ \\
\textit{Bounds} & $R_h\in[0.25,2.5]$ & $0.389{\pm}.135$ & $0.199{\pm}.104$ \\
& $R_h\in[0.1,4]$ & $\mathbf{0.318}{\pm}.013$ & $\mathbf{0.144}{\pm}.014$ \\
& $R_h\in[0.05,6]$ & $0.368{\pm}.088$ & $0.181{\pm}.075$ \\
\textit{Caps} & $c_\varepsilon=c_\eta=0.15$ & $0.662{\pm}.288$ & $0.390{\pm}.202$ \\
& $c_\varepsilon=c_\eta=0.30$ & $\mathbf{0.318}{\pm}.013$ & $\mathbf{0.144}{\pm}.014$ \\
& $c_\varepsilon=c_\eta=0.60$ & $0.439{\pm}.096$ & $0.240{\pm}.073$ \\
\textit{Removal} & No incidence correction & $0.332{\pm}.025$ & $0.156{\pm}.025$ \\
& No observation correction & $0.410{\pm}.153$ & $0.215{\pm}.115$ \\
\bottomrule
\end{tabular}
\captionsetup{skip=2pt}
\captionof{table}{Structural sensitivity on hospital admissions.}
\label{tab:rollout_structure_sensitivity}
\end{center}

\end{document}